\documentclass[letterpaper, 10 pt, conference]{ieeeconf} 
\IEEEoverridecommandlockouts                              % This command is only needed if 
\usepackage{cite}
\usepackage{amsmath,amssymb,amsfonts}
\usepackage{algorithmic}
\usepackage{graphicx}
\usepackage{textcomp}
\usepackage{xcolor}
\usepackage[colorinlistoftodos]{todonotes}
\usepackage{subfigure}
\usepackage{fancyhdr}  
\usepackage{tabularx}
\usepackage{lmodern}   
\usepackage{float}
\usepackage{booktabs}
\usepackage[T1]{fontenc}
\usepackage[utf8]{inputenc}
\usepackage{makecell}
\usepackage{url}

\begin{document}

% working title, can certainly change
\title{Autonomous Reinforcement Learning Robot Control with Intel's Loihi 2 Neuromorphic Hardware\\}

\author{ Kenneth Stewart$^{1*}$, Roxana Leontie$^{1*}$, Samantha Chapin$^{1*}$, Joe Hays$^{1}$, \\
Sumit Bam Shrestha$^{2}$, and Carl Glen Henshaw$^{1}$% <-this % stops a space
\thanks{*These authors contributed equally to this work.}% <-this % stops a space
\thanks{Authors are with the $^{1}$ U.S. Naval Research Laboratory Naval, Washington, DC, USA and $^{2}$ Intel Labs, Santa Clara, USA.
        {\tt\small Corresponding author email: kenneth.m.stewart45.civ@us.navy.mil}}%%
}

\maketitle

\begin{abstract}
We present an end-to-end pipeline for deploying reinforcement learning (RL) trained Artificial Neural Networks (ANNs) on neuromorphic hardware by converting them into spiking Sigma-Delta Neural Networks (SDNNs). We demonstrate that an ANN policy trained entirely in simulation can be transformed into an SDNN compatible with Intel's Loihi 2 architecture, enabling low-latency and energy-efficient inference. As a test case, we use an RL policy for controlling the Astrobee free-flying robot, similar to a previously hardware in space-validated controller. The policy, trained with Rectified Linear Units (ReLUs), is converted to an SDNN and deployed on Intel's Loihi 2, then evaluated in NVIDIA's Omniverse Isaac Lab simulation environment for closed-loop control of Astrobee's motion. We compare execution performance between GPU and Loihi 2. The results highlight the feasibility of using neuromorphic platforms for robotic control and establish a pathway toward energy-efficient, real-time neuromorphic computation in future space and terrestrial robotics applications.
% We have established a pipeline for deploying Reinforcement Learning (RL) trained Artificial Neural Networks (ANNs) converted to spiking Sigma-Delta Neural Networks (SDNNs) in neuromorphic hardware for robotic control. We show that an ANN trained in simulation can be converted to a neuromorphic hardware compatible SDNN and run inference using neuromorphic hardware with lower energy and latency with comparable performance. This experiment's test case is a RL ANN policy trained to control the motion of an Astrobee free-flying robot; a similar policy was previously validated in-space for this robot ~\cite{2025_APIARY_iSpaRo_Paper_1} ~\cite{2025_APIARY_iSpaRo_Paper_2}. The Rectified Linear Units (ReLU) policy is then converted to a SDNN to run on Intel's Loihi 2 neuromorphic hardware, and its performance is validated in tests using the NVIDIA Omniverse Isaac Lab simulation environment to command the Astrobee's motion. The results compare the performance of the policies on the GPU, CPU, and Loihi 2 hardware. This pipeline paves the way for deploying neuromorphic hardware, such as Intel's Loihi, for robotics applications. 
\end{abstract}

\section{Introduction}

Power constraints are a critical consideration for many robotic applications, particularly in space and mobile environments. While data-driven learning on GPUs has pushed substantial progress in robotics, the associated energy demands can hinder their deployment in power-sensitive applications. This paper explores a path toward lower-power robotic control by leveraging neuromorphic hardware. We present a pipeline for training ANNs for robotic control using RL and then converting these ANNs into SDNNs for execution on Intel's Loihi 2 neuromorphic hardware. This approach capitalizes on the ease of training ANNs, and their ease of transition from sim-2-real, while exploiting the energy efficiency of Spiking Neuronal Networks (SNNs). Our approach is validated through simulated control of an Astrobee free-flying robot, within the NVIDIA Omniverse Isaac Lab simulation environment, a task also previously demonstrated on robotic hardware with an embedded CPU in-space~\cite{2025_APIARY_iSpaRo_Paper_1}. This research paves the way for the wider adoption of neuromorphic computing in robotics and opens new avenues for deploying advanced robotics in power-constrained environments.

Beyond small free-flying platforms such as Astrobee, a wide range of real-world missions highlight the need for robust, low-power control systems that can sustain long-duration operation under strict energy budgets. Space exploration is historically constrained with respect to computing power available for a given mission~\cite{grappling_spacecraft}. This is the result of environmental challenges, such as radiation restricting the number of systems rated to survive the desired lifespan of a mission in addition to size, weight, and power, and cost (SWaP-C) requirements for a target mission~\cite{2023_AI_in_space}. Current space-rated radiation-hardened (rad-hard) processors can survive in-space but have limited performance when compared to the state-of-the-art deployed on the ground~\cite{2023_AI_in_space}. 

Several past space and deep-sea robotic platforms ultimately faced limitations driven by actuator degradation, restricted onboard computation, and power exhaustion, conditions under which neuromorphic RL-based control could have provided tangible benefits. For example, NASA's Kepler mission suffered premature degradation of its reaction wheels, resulting in reduced pointing accuracy and eventually the end of its primary mission~\cite{kepler_k2}. More adaptive torque-management policies, learned through reinforcement learning and executed at low-power on neuromorphic hardware, could have mitigated reaction-wheel loading and prolonged operational life. Similarly, autonomous spacecraft such as Deep Space 1 operated with limited onboard computational resources, constraining the frequency and complexity of autonomous control updates~\cite{deep_space_1}. A neuromorphic controller capable of continuous, low-power inference could have enabled more resilient attitude and propulsion control during sensor anomalies without exceeding the spacecraft's stringent power envelope. 

Comparable constraints appear in naval and deep-sea exploration robotics, where battery life and communication limits restrict high levels of onboard autonomy. Vehicles such as WHOI's Nereus~\cite{HROV_nereus} faced navigation drift, energy limitations, and control-induced stress cycles that ultimately contributed to mission loss or shortened deployments. RL-based neuromorphic controllers capable of optimizing thrust usage, compensating for sensor dropout, and minimizing unnecessary control oscillations could have extended endurance while reducing mechanical fatigue. These historical examples underscore that energy-efficient adaptive control is not merely desirable, it is often mission-critical.

Collectively, these considerations motivate the development of control pipelines that combine data-driven learning, fault-tolerant policy adaptation, and low-power neuromorphic execution, as pursued in this work. This is particularly crucial in demanding environments, such as space and naval exploration. By demonstrating an ANN-to-SDNN conversion pipeline for RL-based robotic control and validating it in a high-fidelity simulation environment, this paper capitalizes on the increasing sophistication of RL training algorithms for robust, sim-to-real policies to enable the next generation of autonomous systems capable of long-term, energy-efficient operation, ultimately addressing these resource constraints with platforms like Loihi 2.

\section{Related Work}

The development of robust autonomous robotic systems has been significantly advanced by training ANNs on large amounts of data~\cite{embodimentcollaboration2025openxembodimentroboticlearning, schulman2017ppo}. Typical Deep Reinforcement Learning (DRL) algorithms require broad exploration of large state spaces, which in turn needs large volume of interaction data, often difficult to collect in the real world~\cite{schulman2017ppo, openai2019dota2largescale}. Training SNNs with binary spikes to achieve accuracies comparable to ANNs remains a significant challenge, partly due to the need to backpropagate loss through time~\cite{lee2016trainingdeepspikingneural}. Training an SNN requires roughly 10x more time than training an ANN with the same architecture~\cite{brehove2025sigmadeltaneuralnetworkconversion}. Prior approaches to ANN-to-SNN conversion have primarily relied on encoding activations as spike rates~\cite{rueckauer2017conversiondnntoevent}, which necessitates running numerous simulation timesteps per input to accurately estimate output spike rates. However, Loihi 2's support for quantized graded spikes enables a new pathway: converting a trained ANN into a quantized graded-spike SNN that leverages both temporal and spatial sparsity.

Loihi 2 is Intel's second-generation neuromorphic research processor~\cite{orchard2021loihi2}. It features a fully event-driven, digitally asynchronous network of neuromorphic cores that support diverse synaptic interconnects, including convolutional connectivity, along with fully programmable spiking neuron dynamics and integer-valued (graded) spikes. Designed to exploit the sparse communication and event-driven computation patterns characteristic of SNNs, Loihi 2 enables low-power and low-latency processing with sparse architectures. This inherent compatibility with sparse, event-driven computation makes Loihi 2 a particularly well-suited platform for implementing Sigma-Delta Neural Networks (SDNNs).

SDNNs minimize redundant computations by transmitting only activation changes~\cite{oconnor2016sigmadeltaquantizednetworks}. This temporal difference encoding achieves computational efficiency without compromising model accuracy. Building on this concept,~\cite{shrestha2023efficientvideoaudioprocessing} implemented an SDNN on Intel's Loihi 2 and compared its performance to that of an NVIDIA Jetson platform. Their results highlighted the benefits of neuromorphic hardware for SDNNs, leveraging Loihi 2's event-driven processing capabilities. The works of~\cite{shrestha2023efficientvideoaudioprocessing} and~\cite{sarkar2025yolokp} importantly demonstrated how a trained ANN can be successfully converted to an SDNN and run on the Intel Loihi 2, giving the benefits of ANN training and accuracy with the lower energy and sparsity benefits of running SDNNs on neuromorphic hardware. While this has been demonstrated for audio and visual signal processing using ANN trained networks converted to SDNNs has not yet been demonstrated for robotic control. 

Robotic control has previously been demonstrated on Intel's Loihi 1 neuromorphic hardware~\cite{dewolf2023robotcontrol}. In~\cite{dewolf2023robotcontrol}, a simulated Kinova Generation 3 7-Degree-of-Freedom (DOF) arm was controlled to perform a reaching task. A rate-based network was converted into an SNN, with the original network trained under activation noise to increase robustness to the additional variability introduced during SNN inference. The network was further adapted to satisfy Loihi 1's hardware constraints. The same approach was then used to train the system on two manipulation tasks: opening a cabinet door and a drawer, with the physical robotic arm operating in a kitchen environment. The torque control commands were generated by the network running directly on Intel's Loihi 1. Video results can be seen in~\cite{leontie2023edgeintelligent}.

While these works demonstrate the feasibility of deploying SNN-based controllers on neuromorphic hardware, they rely on converting rate-based controllers optimized specifically for Loihi's constraints. Closer to our work,~\cite{zanatta2024exploring} explore a different direction by training  Proximal Policy Optimization (PPO) based SNN policies directly within Isaac Gym, enabling rapid exploration of SNN configurations for deep reinforcement learning robotic tasks. However, only very shallow networks of a single hidden layer had reasonable performance compared to ANNs or other SNN training methods and the networks were not deployed on neuromorphic hardware. In contrast, our approach focuses on converting a PPO-trained ANN policy into an SNN suitable for execution on neuromorphic hardware that is scalable to deeper networks, thus leveraging the stability and performance of conventional ANN training while producing a spiking policy ready for efficient deployment. 

Demonstrating this capability, ANN to SDNN converted networks for robotic control on neuromorphic hardware, represents a critical step toward unlocking the true potential of neuromorphic robotics. We showcase this conversion for an ANN policy to control of a 6-DOF zero-gravity (zero-G) free-flyer in simulation, detailed further in the next section.

\subsection{Example Task Background: 6 Degrees of Freedom Robot Control}

The example RL task tested in this paper was initially tested as part of the Autonomous Planning In-space Assembly Reinforcement-learning free-flYer (APIARY) experiment~\cite{2025_APIARY_iSpaRo_Paper_1}. This experiment was tested on the NASA Astrobee on-board the International Space Station to control the 6-DOF motion of the free-flying robot in zero-G. APIARY used the NVIDIA Issac Lab simulation environment to train an actor-critic PPO network. More detailed information about the curriculum training~\cite{2025_APIARY_iSpaRo_Paper_2} and ground/space testing\cite{2025_APIARY_iSpaRo_Paper_1} can be found in the associated papers. It's important to note that, for the present investigation, the RL policy required re-training specifically using Rectified Linear Units (ReLU) to ensure compatibility with the baseline ANN to SDNN conversion~\cite{shrestha2023efficientvideoaudioprocessing}, detailed in the \textit{Methods} section. Due to the inherent limitations of the Astrobee platform in its current configuration, namely the absence of an on-board neuromorphic processor, this paper's primary emphasis is centered on a detailed analysis of the Loihi's performance in controlling a simulated Astrobee within the NVIDIA Isaac Lab environment. To ensure a valid and meaningful assessment, we replicated the training setup and the learning curriculum that were originally utilized for the flight policy, thereby enabling a direct and equitable comparison of performance metrics. Ultimately, our broader objective is to demonstrate the feasibility and potential of enabling neuromorphic-based control for space robots. The imperative to reduce the size, weight, and power consumption (SWaP) of hardware deployed in space operations stems from the historically significant costs associated with launching mass and volume into orbit, coupled with the inherent power and data resource constraints that govern continuous operations in this environment.

\subsection{Methods}

Nvidia's Omniverse Isaac Lab~\cite{mittal2023orbit} is a real-time high fidelity physics simulator that is usable for robotics applications with Isaac Lab~\cite{naveed2024omniverse_survey}. The sim can support the parallelized RL training on the order of 10000s of robots at a time for highly efficient training. The simulator can be used to run robotic simulation environments in zero-G which is ideal for space robotics application development. As in previous work~\cite{2025_APIARY_iSpaRo_Paper_1, 2025_APIARY_iSpaRo_Paper_2} an Astrobee robot is trained to move to goal positions and orientations in zero-G by outputting the appropriate force and torque for movement. A 12x64x64x6 layer Actor in an Actor-Critic network was trained with PPO~\cite{schulman2017ppo} using a clipped objective and Generalized Advantage Estimation (GAE)~\cite{schulman2018gae}:

\begin{equation}
\begin{aligned}
L^{\mathrm{CLIP}}(\theta)
= \hat{\mathbb{E}}_t \biggl[
\min\Bigl(
& \rho_t(\theta)\,\hat{A}_t,\;\\
& \operatorname{clip}\bigl(\rho_t(\theta),\,1-\epsilon,\,1+\epsilon\bigr)\,\hat{A}_t
\Bigr)
\biggr]
\end{aligned}
\end{equation} 

where $\rho_t(\theta) := \frac{\pi_\theta(a_t\mid s_t)}{\pi_{\theta_{\text{old}}}(a_t\mid s_t)}$ and $\pi_{\theta}(a_t|s_t)$ is the probability of taking action $a_t$ in state $s_t$ under the current policy with parameters $\theta$, with clipping at value $\epsilon$ and $\hat{A}_t$ is the advantage estimate computed using GAE as: 

\begin{equation}
	\delta_t = r_t + \gamma V(s_{t+1}) - V(s_t)
\end{equation} 

\begin{equation}
	\hat{A}_t^{\text{GAE}(\gamma,\lambda)} = \sum_{l=0}^{\infty} (\gamma\lambda)^l \delta_{t+l}
\end{equation}

Where $\delta_t$ is the Temporal Difference (TD) residual, $V(s_t)$ is the estimated value function at state $s_t$, $r_t$ is the immediate reward at time step $t$, $\gamma$ is the discount factor for future rewards, and $\lambda$ is the GAE smoothing parameter. 

Unlike~\cite{2025_APIARY_iSpaRo_Paper_2} that uses Exponential Linear Unit (ELU) activations, the SDNN network uses ReLU activations to be compatible with SDNN conversion for the Loihi 2. The input layer is 12 representing observations given by linear velocity, angular velocity, position error relative to the goal position, and orientation error relative to the goal orientation. Since only the Actor network is needed during inference the Actor is converted to an SDNN. The input layer is converted to a Delta layer, the hidden layers are converted to Sigma Delta ReLU layers, and the output layer is converted to a Sigma layer. 

The Delta layer uses a delta encoder, an input difference operator, that sends an output spike to the next layer only if the information exceeds its last communicated reference value as follows~\cite{shrestha2023efficientvideoaudioprocessing}: 

\begin{equation}
\begin{split}
	s[t] = & (x[t]-x_{ref}[t-1]\mathcal{H}(|x[t]-x_{ref}[t-1])-\vartheta) \\
        x_{ref}[t] = & x_{ref}[t-1] + s[t]
\end{split}
\end{equation}

Where $x_{ref}[t]$ is the last communicated reference value, $\vartheta$ is the threshold, and $\mathcal{H}$ is the heaviside step function~\cite{shrestha2023efficientvideoaudioprocessing}. The Sigma-Delta-ReLU layers are a Sigma and a Delta encapsulated by a ReLU activation function. 

% The Sigma-Delta pair between each layer sparsifies spike messages, with delta encoding logic sending a spike only when a large enough change crosses a threshold~\cite{sarkar2025yolokp}. 

The Sigma decoder on the receiving end of the Delta output performs a summation of the incoming signal to reconstruct the original input~\cite{shrestha2023efficientvideoaudioprocessing}:

\begin{equation}
        x_{rec}[t] = x_{rec}[t-1] + s[t]
\end{equation} 

where xrec is the received signal. Applying Sigma-Delta encapsulation to the ReLU activation in ANNs provides a simple yet effective method for enabling efficient sparse computation within standard ANN frameworks through the use of Sigma-Delta-ReLU layers. 

The SDNN layers have an activation threshold of 0.1, which is the threshold level for changes to be passed to the next layer with higher values yielding more sparsity. If the threshold is 0 the cumulative output equals the network. Loihi 2 supports graded spikes where a spike can carry up to 24 bits of integer magnitude. The SDNN uses graded spikes. Loihi 2's neurocores only allow integer operations, so the SDNN is quantized and when running on the Loihi 2 the computational graph of the integer-based SDNN is mapped onto the Loihi 2 chip's neurocores using NxKernel, Intel's proprietary software stack~\cite{sarkar2025yolokp}. 

Observations taken from the Isaac-Sim simulation at each time step are floating point values. To run on the SDNN, and the Loihi 2, the observations must be quantized. Output actions are dequantized back into floating precision values for input into the simulation environment to advance the simulation to the next step.
%\todo[inline]{Explain Output actions are dequantized back into floating precision values for input into the simulation environment to advance the simulation to the next step}
%\todo[inline]{Explain differences between cpu/gpu and loihi for this for some reason} 

The Loihi 2 board used in experiments is an Intel Kapoho Point N3C1 board with 8 Loihi 2 chips. A super host machine executes the simulation code, loads the SDNN, maps the SDNN onto the board, and communicates with the Loihi 2 over ssh to run the SDNN on chip and obtain energy measurements while running. The SDNN network fits within 1 of the 8 chips. Experiments and comparisons are made with the ANN running on GPU, and the SDNN running on the Loihi 2. 

\subsection{Results}

\begin{figure}[ht]
    \centering 
    \includegraphics[width=\linewidth]{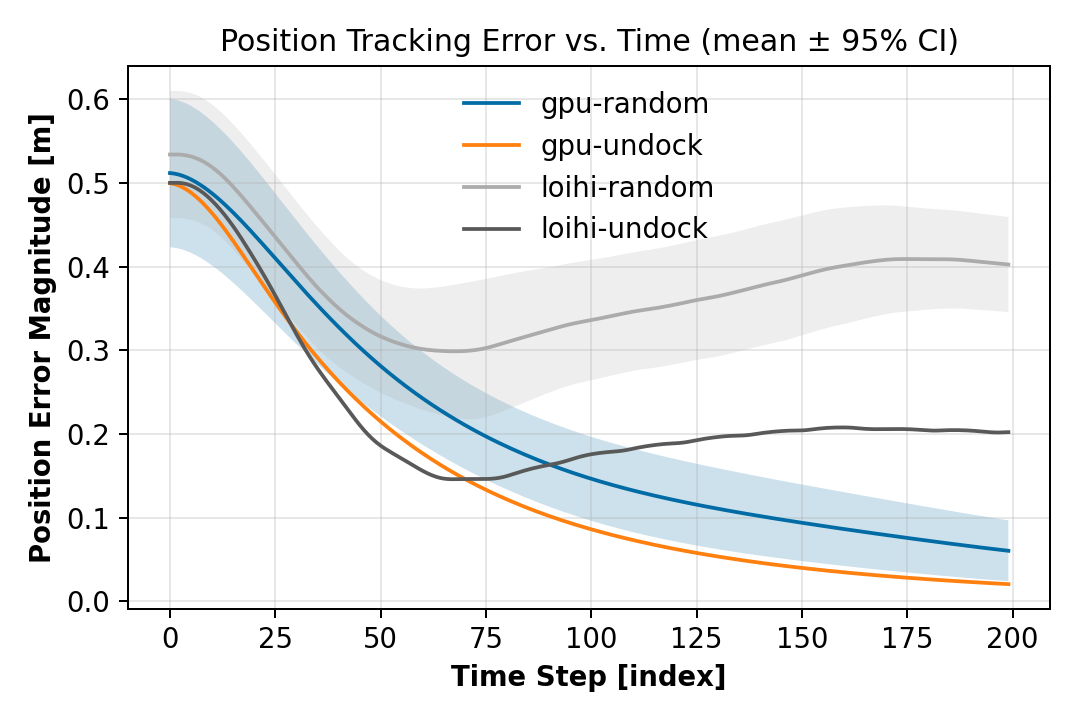}
    \caption{Averaged position tracking error, in meters (m), for data sets of 10 random and 10 undock (0.5-meter X-axis) movements comparing the performance of the GPU (blue and orange lines) and Loihi 2 (light grey and dark grey). The mean is represented by the solid lines with the +- 95\% confidence interval (CI) shaded in the respective colors.}
    \label{fig:fig_position_error_timeseries} 
\end{figure}

\begin{figure}[ht]
    \centering 
    \includegraphics[width=\linewidth]{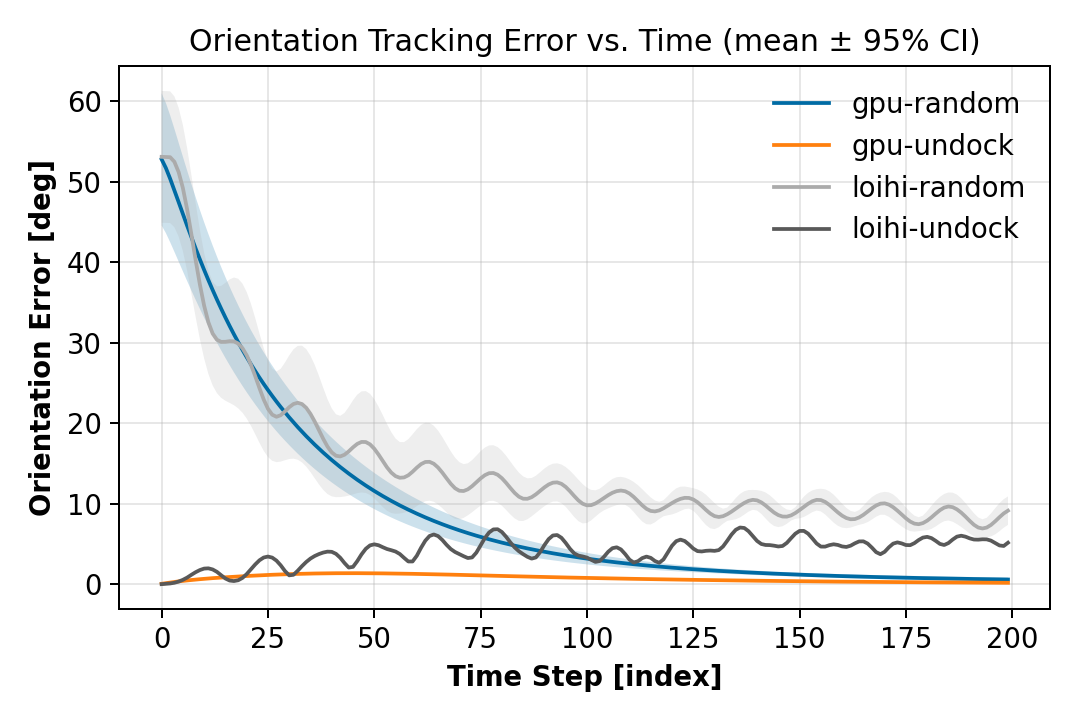}
    \caption{Averaged orientation tracking error, in degrees (deg), for data sets of 10 random and 10 undock (0.5-meter X-axis) movements comparing the performance of the GPU (blue and orange lines) and Loihi 2 (light grey and dark grey). The mean is represented by the solid lines with the +- 95\% confidence interval (CI) shaded in the respective colors.}
    \label{fig:fig_orientation_error_timeseries_deg} 
\end{figure}

Experiments were done in NVIDIA's Isaac Lab Omniverse simulation environment~\cite{mittal2023orbit}. An undock maneuver, the same maneuver tested on the ISS in hardware and evaluated here~\cite{2025_APIARY_iSpaRo_Paper_1}, which is the equivalent to moving 0.5 meters along the X axis while maintaining orientation and positions along the y and z axis, was performed 10 times over 10 seeds. To prove the capability of the policy to perform more complex maneuvers, 10 randomized maneuvers were performed with varying goal position and orientation, The random goal position was in the range (x=+/-0.5m,y=+/-0.5m,z=+/-0.5m) meters from the initial position and the goal orientation was a quaternion within (w=1.0, x=+/-0.5,y=+/0.5,z=+/0.5) which in Euler angles is approximately +/-60 degrees (or 0-120 degrees) along each axis from the initial orientation. Experiments are evaluated across 200 simulation time steps, which is enough time for maneuvers to be completed. The results of the position and orientation tracking error averaged across the 10 random and undock trials are plotted in Figure~\ref{fig:fig_position_error_timeseries} and~\ref{fig:fig_orientation_error_timeseries_deg} respectively. 

\begin{figure}[ht]
    \centering 
    \includegraphics[width=\linewidth]{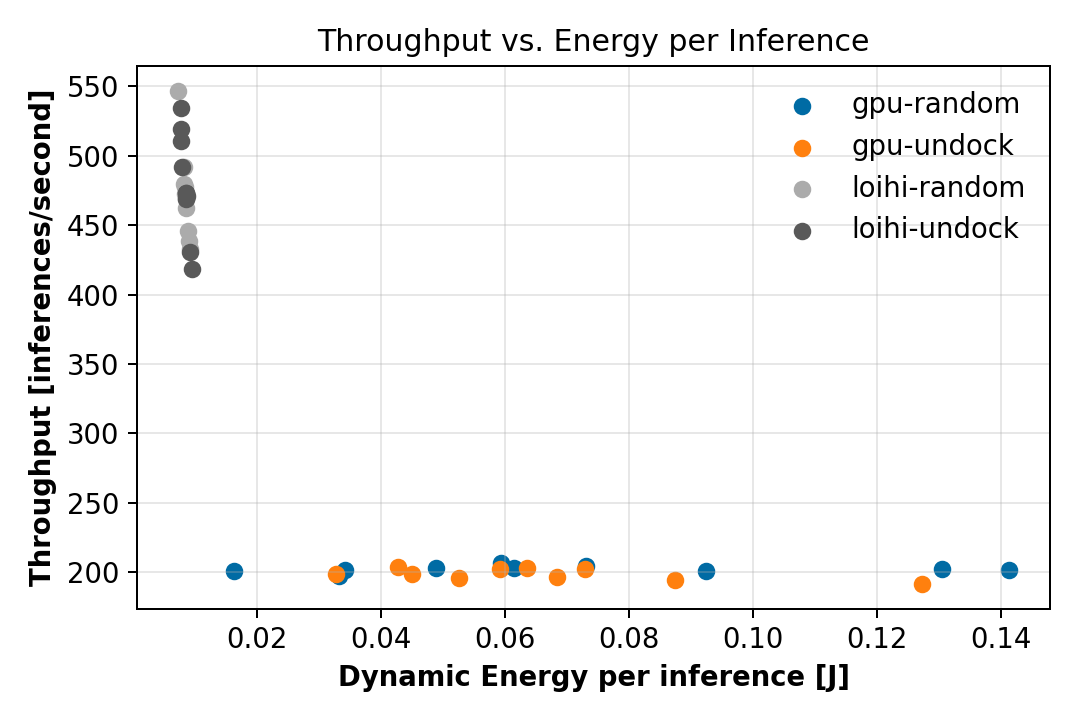}
    \caption{Plotting the throughput per inference, in seconds, vs. energy per inference, in joules (J), for the 10 random and 10 undock (0.5-meter X-axis) movements comparing the performance of the GPU (blue and orange lines) and Loihi 2 (light grey and dark grey).}
    \label{fig:fig_throughput_vs_energy_per_inf} 
\end{figure}

The maneuvers were evaluated running with Ubuntu 20.04 LTS and Nvidia's Isaac-Sim version 5.0.0 and Isaac Lab version 2.2.0~\cite{githubReleaseV220} on a Nvidia Quadro RTX 8000 GPU and an Intel Loihi 2 N3C1 neuromorphic processor running Lava-dl version 0.5.0~\cite{githubGitHubBamsumitlavadl}, Lava version 0.9.0~\cite{githubReleaseLava}, and nxkernel version 0.4.1. The GPU was chosen because currently Nvidia's Isaac-Sim requires an RTX GPU to run so comparisons to Jetson class boards cannot be made for these experiments. Latency and energy for the GPU is measured using the Python ZeusMonitor library~\cite{githubReleaseZeus} designed for deep learning workloads at inference time. Throughput is reported as the number of inferences the network and the hardware running on it can do per second with Figure~\ref{fig:fig_throughput_vs_energy_per_inf} showing the throughput vs Energy per inference for the random and undock movements comparing ANN on GPU, and SDNN on Loihi 2.

% \begin{table*}[H]
%         \caption{Network performance comparison (mean $\pm$ SD)}
%         \label{tab:summary_mean_sd}
%         \centering
%         \resizebox{\textwidth}{!}
%                 {\begin{tabular}{llrrrrrrrrr}
%                 \toprule
%                 Platform & Task & RMSE Pos [m] & Final Pos [m] & RMSE Ori [deg] & Final Ori [deg] & Total Energy/inf [J] & Dynamic Energy/inf [J] & Latency/inf [ms] & Throughput [inf/s] & EDP/inf [mJs] \\
%                 \midrule
%                 GPU & random & 0.142 ± 0.048 & 0.060 ± 0.059 & 15.468 ± 3.721 & 0.581 ± 0.397 & 0.217 ± 0.041 & 0.069 ± 0.041 & 4.944 ± 0.062 & 202.283 ± 2.546 & 1.074 ± 0.203 \\
%                 LOIHI & random & 0.225 ± 0.047 & 0.402 ± 0.092 & 18.198 ± 4.882 & 9.136 ± 2.852 & 0.013 ± 0.001 & 0.008 ± 0.001 & 4.257 ± 0.277 & 471.675 ± 32.492 & 55542.014 ± 7034.641 \\
%                 GPU & undock & 0.118 & 0.021 & 0.829 & 0.171 & 0.216 ± 0.029 & 0.065 ± 0.027 & 5.034 ± 0.103 & 198.725 ± 4.052 & 1.090 ± 0.166 \\
%                 LOIHI & undock & 0.143 & 0.202 & 4.399 & 5.152 & 0.013 ± 0.001 & 0.008 ± 0.001 & 4.197 ± 0.328 & 479.149 ± 36.602 & 54069.424 ± 8364.472 \\
%                 \bottomrule
%                 \end{tabular}}
% \end{table*}

\begin{table*}[t]
\caption{Network performance goal comparison (mean $\pm$ SD)}
\label{tab:summary_pose_mean_sd}
\centering
% \resizebox{\columnwidth}{!}{
    \begin{tabular}{llrrrr}
    \toprule
    Platform & Task & \makecell{RMSE \\ Position $[m]$} & \makecell{Final \\ Position $[m]$} & \makecell{RMSE \\ Orientation $[deg]$} & \makecell{Final \\ Orientation $[deg]$} \\
    \midrule
    GPU & random & 0.142 ± 0.048 & 0.060 ± 0.059 & 15.468 ± 3.721 & 0.581 ± 0.397 \\
    LOIHI & random & 0.225 ± 0.047 & 0.402 ± 0.092 & 18.198 ± 4.882 & 9.136 ± 2.852 \\
    GPU & undock & 0.118 ± 0.000 & 0.021 ± 0.000 & 0.829 ± 0.000 & 0.171 ± 0.000 \\
    LOIHI & undock & 0.143 ± 0.000 & 0.202 ± 0.000 & 4.399 ± 0.000 & 5.152 ± 0.000 \\
    \bottomrule
    \end{tabular}
  % }
\end{table*}

\begin{table*}[t]
\caption{Network performance comparison (mean $\pm$ SD)}
\label{tab:summary_energy_mean_sd}
\centering
% \resizebox{\columnwidth}{!}{
  \begin{tabular}{llrrrrr}
  \toprule
  Platform & Task & \makecell{Total \\ Energy/inf $[J]$} & \makecell{Dynamic \\ Energy/inf $[J]$} & \makecell{Latency/inf \\ $[ms]$} & \makecell{Throughput \\ $[inf/s]$} & \makecell{EDP/inf \\ $[mJs]$} \\
  \midrule
  GPU & random & 0.217 ± 0.041 & 0.069 ± 0.041 & 4.944 ± 0.062 & 202.283 ± 2.546 & 1.074 ± 0.203 \\
  LOIHI & random & 0.013 ± 0.001 & 0.008 ± 0.001 & 4.257 ± 0.277 & 471.675 ± 32.492 & 0.056 ± 0.007 \\
  GPU & undock & 0.216 ± 0.029 & 0.065 ± 0.027 & 5.034 ± 0.103 & 198.725 ± 4.052 & 1.090 ± 0.166 \\
  LOIHI & undock & 0.013 ± 0.001 & 0.008 ± 0.001 & 4.197 ± 0.328 & 479.149 ± 36.602 & 0.054 ± 0.008 \\
  \bottomrule
  \end{tabular}
  % }
\end{table*}

\begin{figure}[ht]
        \centering 
        \begin{tabularx}{\textwidth}{cc}
                (a) \includegraphics[height=3cm]{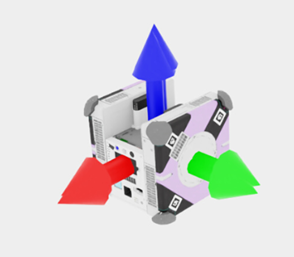} &%
                (b) \includegraphics[height=3cm]{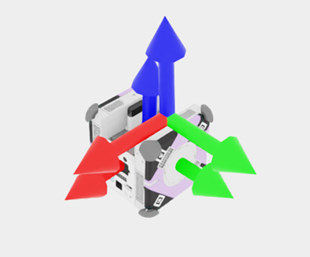}
        \end{tabularx}
        \caption{(a) ANN GPU inference (b) SDNN Loihi inference. \\ If the Astrobee robot and the goal position and orientation arrows overlap the goal is precisely reached. This shows an example output of the undock task. }
        \label{fig:GPUvsLoihiAstrobeePic}
\end{figure}

Table~\ref{tab:summary_pose_mean_sd} shows results comparing the performance of the ANN running on GPU to the SDNN running on Loihi 2 in terms of task position and orientation error on maneuvering towards a random goal and the fixed undock goal, and Table~\ref{tab:summary_energy_mean_sd} compares the energy, latency, throughput, and EDP of the same networks, hardware, and tasks per inference. Since spikes are passed from one layer to the next during each time step and the network has 2 hidden layers, the latency of the SDNN on Loihi was reported as 2x the average measured time per time step of execution. All results in the tables show the mean +/- 1 standard deviation of 10 runs over 10 different seeds. In Table~\ref{tab:summary_pose_mean_sd} the position and orientations are given as averages across all axis of motion. The SDNN does not achieve the same the same task performance as the ANN, particularly when maneuvering to random positions. The ANN on GPU has 0.083m less RMSE position error and 8.55 degrees less RMSE orientation error over the movement trajectory towards the goal than the SDNN on the random movement task, and 0.025m less RMSE position error on the undock task. The final position and orientation error for the SDNN on Loihi compared to the RMSE indicates the network starts on a good trajectory towards the goal but likely accumulation of errors from the spike encoding and quantization needed to run on the Loihi 2 gives a worse final position and orientation relative to the goal, which can be seen in Figures~\ref{fig:fig_position_error_timeseries} and~\ref{fig:fig_orientation_error_timeseries_deg}. A visualization of this ending position is shown in Figure~\ref{fig:GPUvsLoihiAstrobeePic}, shows that while the SDNN does not move as precisely to the goal orientation and position the SDNN does get close and does not greatly deviate from moving to the desired goal and is stable at the final position and orientation. As shown in Table~\ref{tab:summary_energy_mean_sd}, the SDNN on Loihi 2 is far more energy efficient with lower latency and higher throughput. The SDNN on Loihi 2 has ~5\% of the EDP of the ANN on GPU, meaning the SDNN on Loihi 2 is about 20x more energy efficient and has more than 2x throughput than the ANN on GPU. The Dynamic Energy, the approximate energy usage of just the neural network operations during inference, is also lower for the SDNN on Loihi with the SDNN using only ~12.3\% of the Dynamic Energy of the ANN on GPU. There is a clear performance vs energy usage tradeoff between using the ANN on GPU and SDNN on the Loihi 2. There are ways that the SDNN performance can be made closer to the ANN results in future work. For example, an untrained Delta encoding network converts the fp32 observation inputs to spikes that are then quantized to int24 representation for processing by the SDNN on Loihi 2. Performance could be improved by training the network to better represent the observation data input to the SDNN. Greater exploration of hyperparameters that give better performance for the SDNN converted network could also enhance the performance of the SDNN on Loihi 2 while preserving lower energy consumption compared to the ANN on GPU. 

\subsection{Conclusion}

Overall, this initial experiment demonstrates the feasibility of executing a RL-based robot control policy with neuromorphic hardware. Depending on the target application, the observed performance deviation could be acceptable when weighting against the other tradeoffs in latency, energy, and performance. Future work may further mitigate these limitations, as described in the results section. Although the control network used here is relatively small, and has been previously deployed on a CPU during our previous Astrobee ISS operations, we anticipate that larger and more complex policies will exhibit different SWaP requirements, making the advantages of neuromorphic execution even more significant. 

\section*{Acknowledgment}
Thanks to ONR for supporting our research. Thanks to Intel Labs for providing their hardware and support.

\bibliographystyle{IEEEtran}
\bibliography{example} % .bib 

@misc{githubReleaseV220,
	author = {},
	title = {{R}elease v2.2.0 · isaac-sim/{I}saac{L}ab --- github.com},
	howpublished = {\url{https://github.com/isaac-sim/IsaacLab/releases/tag/v2.2.0}},
	year = {},
	note = {[Accessed 14-11-2025]},
}

@misc{githubReleaseLava,
	author = {},
	title = {{R}elease {L}ava 0.9.0 · lava-nc/lava --- github.com},
	howpublished = {\url{https://github.com/lava-nc/lava/releases/tag/v0.9.0}},
	year = {},
	note = {[Accessed 14-11-2025]},
}

@misc{githubGitHubBamsumitlavadl,
	author = {},
	title = {{G}it{H}ub - bamsumit/lava-dl at ann\_sdnn --- github.com},
	howpublished = {\url{https://github.com/bamsumit/lava-dl/tree/ann_sdnn}},
	year = {},
	note = {[Accessed 14-11-2025]},
}

@misc{githubReleaseZeus,
	author = {},
	title = {{R}elease {Z}eus v0.13.0 · ml-energy/zeus --- github.com},
	howpublished = {\url{https://github.com/ml-energy/zeus/releases/tag/zeus-v0.13.0}},
	year = {},
	note = {[Accessed 14-11-2025]},
}

@ARTICLE{rueckauer2017conversiondnntoevent,
  
AUTHOR={Rueckauer, Bodo  and Lungu, Iulia-Alexandra  and Hu, Yuhuang  and Pfeiffer, Michael  and Liu, Shih-Chii },
         
TITLE={Conversion of Continuous-Valued Deep Networks to Efficient Event-Driven Networks for Image Classification},
        
JOURNAL={Frontiers in Neuroscience},
        
VOLUME={Volume 11 - 2017},

YEAR={2017},

URL={https://www.frontiersin.org/journals/neuroscience/articles/10.3389/fnins.2017.00682},

DOI={10.3389/fnins.2017.00682},

ISSN={1662-453X},

}

@misc{lee2016trainingdeepspikingneural,
      title={Training Deep Spiking Neural Networks using Backpropagation}, 
      author={Jun Haeng Lee and Tobi Delbruck and Michael Pfeiffer},
      year={2016},
      eprint={1608.08782},
      archivePrefix={arXiv},
      primaryClass={cs.NE},
      url={https://arxiv.org/abs/1608.08782}, 
}

@misc{openai2019dota2largescale,
      title={Dota 2 with Large Scale Deep Reinforcement Learning}, 
      author={OpenAI and Christopher Berner and Greg Brockman and Brooke Chan and Vicki Cheung and Przemys{\l{}}aw D{\k{e}}biak and Christy Dennison and David Farhi and Quirin Fischer and Shariq Hashme and Chris Hesse and Rafal Józefowicz and Scott Gray and Catherine Olsson and Jakub Pachocki and Michael Petrov and Henrique P. d. O. Pinto and Jonathan Raiman and Tim Salimans and Jeremy Schlatter and Jonas Schneider and Szymon Sidor and Ilya Sutskever and Jie Tang and Filip Wolski and Susan Zhang},
      year={2019},
      eprint={1912.06680},
      archivePrefix={arXiv},
      primaryClass={cs.LG},
      url={https://arxiv.org/abs/1912.06680}, 
}

@misc{embodimentcollaboration2025openxembodimentroboticlearning,
  title={Open {X-E}mbodiment: Robotic Learning Datasets and {RT-X} Models},
  author = {{Open X-Embodiment Collaboration} and Abby O'Neill and others},
  howpublished  = {\url{https://arxiv.org/abs/2310.08864}},
  year = {2023},
}

@misc{oconnor2016sigmadeltaquantizednetworks,
      title={Sigma Delta Quantized Networks}, 
      author={Peter O'Connor and Max Welling},
      year={2016},
      eprint={1611.02024},
      archivePrefix={arXiv},
      primaryClass={cs.NE},
      url={https://arxiv.org/abs/1611.02024}, 
}

@misc{shrestha2023efficientvideoaudioprocessing,
      title={Efficient Video and Audio processing with Loihi 2}, 
      author={Sumit Bam Shrestha and Jonathan Timcheck and Paxon Frady and Leobardo Campos-Macias and Mike Davies},
      year={2023},
      eprint={2310.03251},
      archivePrefix={arXiv},
      primaryClass={cs.NE},
      url={https://arxiv.org/abs/2310.03251}, 
}

@misc{sarkar2025yolokp,
      title={Region Masking to Accelerate Video Processing on Neuromorphic Hardware}, 
      author={Sreetama Sarkar and Sumit Bam Shrestha and Yue Che and Leobardo Campos-Macias and Gourav Datta and Peter A. Beerel},
      year={2025},
      eprint={2503.16775},
      archivePrefix={arXiv},
      primaryClass={cs.CV},
      url={https://arxiv.org/abs/2503.16775}, 
}

@misc{brehove2025sigmadeltaneuralnetworkconversion,
      title={Sigma-Delta Neural Network Conversion on Loihi 2}, 
      author={Matthew Brehove and Sadia Anjum Tumpa and Espoir Kyubwa and Naresh Menon and Vijaykrishnan Narayanan},
      year={2025},
      eprint={2505.06417},
      archivePrefix={arXiv},
      primaryClass={cs.NE},
      url={https://arxiv.org/abs/2505.06417}, 
}

@article{dewolf2023robotcontrol,
doi = {10.1088/2634-4386/acb286},
url = {https://doi.org/10.1088/2634-4386/acb286},
year = {2023},
month = {feb},
publisher = {IOP Publishing},
volume = {3},
number = {1},
pages = {014007},
author = {DeWolf, Travis and Patel, Kinjal and Jaworski, Pawel and Leontie, Roxana and Hays, Joe and Eliasmith, Chris},
title = {Neuromorphic control of a simulated 7-DOF arm using Loihi},
journal = {Neuromorphic Computing and Engineering}
}

@article{orchard2021loihi2,
  author       = {Garrick Orchard and
                  Edward Paxon Frady and
                  Daniel Ben Dayan Rubin and
                  Sophia Sanborn and
                  Sumit Bam Shrestha and
                  Friedrich T. Sommer and
                  Mike Davies},
  title        = {Efficient Neuromorphic Signal Processing with Loihi 2},
  journal      = {CoRR},
  volume       = {abs/2111.03746},
  year         = {2021},
  url          = {https://arxiv.org/abs/2111.03746},
  eprinttype    = {arXiv},
  eprint       = {2111.03746},
  bibsource    = {dblp computer science bibliography, https://dblp.org}
}

@article{naveed2024omniverse_survey,
  author = {Ahmed, Naveed and Afyouni, Imad and Dabool, Hamzah and Al Aghbari, Zaher},
  title = {A systemic survey of the Omniverse platform and its applications in data generation, simulation and metaverse},
  journal = {Frontiers in Computer Science},
  volume = {6},
  year = {2024},
  doi = {10.3389/fcomp.2024.1423129},
  url = {https://www.frontiersin.org/articles/10.3389/fcomp.2024.1423129/full},
  issn = {2624-9898}
}

@article{mittal2023orbit,
  author = {Mittal, Mayank and Yu, Calvin and Yu, Qinxi and Liu, Jingzhou and Rudin, Nikita and Hoeller, David and Yuan, Jia Lin and Singh, Ritvik and Guo, Yunrong and Mazhar, Hammad and Mandlekar, Ajay and Babich, Buck and State, Gavriel and Hutter, Marco and Garg, Animesh},
  title = {Orbit: A Unified Simulation Framework for Interactive Robot Learning Environments},
  journal = {IEEE Robotics and Automation Letters},
  year = {2023},
  volume = {8},
  number = {6},
  pages = {3740--3747},
  doi = {10.1109/LRA.2023.3270034}
}

@misc{schulman2017ppo,
  author = {John Schulman and Filip Wolski and Prafulla Dhariwal and Alec Radford and Oleg Klimov},
  title = {Proximal Policy Optimization Algorithms},
  year = {2017},
  eprint = {1707.06347},
  archivePrefix = {arXiv},
  primaryClass = {cs.LG},
  url = {https://arxiv.org/abs/1707.06347}
}

@misc{schulman2018gae,
  author = {John Schulman and Philipp Moritz and Sergey Levine and Michael Jordan and Pieter Abbeel},
  title = {High-Dimensional Continuous Control Using Generalized Advantage Estimation},
  year = {2018},
  eprint = {1506.02438},
  archivePrefix = {arXiv},
  primaryClass = {cs.LG},
  url = {https://arxiv.org/abs/1506.02438}
}

@inproceedings{2025_APIARY_iSpaRo_Paper_1,
  author = {Samantha Chapin and Kenneth Stewart and Roxana Leontie and Carl Glen Henshaw},
  title = {Autonomous Planning In-space Assembly Reinforcement-learning free-flYer (APIARY) International Space Station Astrobee Testing},
  booktitle = {iSpaRo Conference (Accepted)},
  year = {2025},
}

@inproceedings{2025_APIARY_iSpaRo_Paper_2,
  author = {Kenneth Stewart and Samantha Chapin and Roxana Leontie and Carl Glen Henshaw},
  title = {Crossing the Sim2Real Gap Between Simulation and Ground Testing to Space Deployment of Autonomous Free-flyer Control},
  booktitle = {iSpaRo Conference (Accepted)},
  year = {2025},
}

@article{zanatta2024exploring,
  title={Exploring spiking neural networks for deep reinforcement learning in robotic tasks},
  author={Zanatta, Luca and Barchi, Francesco and Manoni, Simone and Tolu, Silvia and Bartolini, Andrea and Acquaviva, Andrea},
  journal={Scientific Reports},
  volume={14},
  number={1},
  pages={30648},
  year={2024},
  publisher={Nature Publishing Group UK London}
}

@misc{leontie2023edgeintelligent,
  author       = {Leontie, Roxana and Hays, Joseph and DeWolf, Travis},
  title        = {Edge Intelligence Enabled Autonomous Systems},
  howpublished = {Presentation, ICONS 2023, Santa Fe, New Mexico},
  year         = {2023},
  note         = {Presented by Joe Hays. Slides 40--41},
  url          = {https://docs.google.com/presentation/d/17mR33Tpl_dMiDm5L_lmoeVXHD8OPLte1/edit?usp=drive_link&ouid=107661738082571430949&rtpof=true&sd=true},
}

@article{grappling_spacecraft,
   author = "Henshaw, Carl Glen and Glassner, Samantha and Naasz, Bo and Roberts, Brian",
   title = "Grappling Spacecraft", 
   journal= "Annual Review of Control, Robotics, and Autonomous Systems",
   year = "2022",
   volume = "5",
   number = "Volume 5, 2022",
   pages = "137-159",
   doi = "https://doi.org/10.1146/annurev-control-042920-011106",
   url = "https://www.annualreviews.org/content/journals/10.1146/annurev-control-042920-011106",
   publisher = "Annual Reviews",
   issn = "2573-5144",
   type = "Journal Article",
}

@misc{2023_AI_in_space,
  author       = {Goodwill, Justin and Wilson, Christopher and MacKinnon, James},
  title        = {Current AI Technology in Space},
  howpublished = {NASA Goddard Space Flight Center},
  year         = {2023},
  note         = {v4},
  url          = {https://ntrs.nasa.gov/api/citations/20240001139/downloads/Current%20Technology%20in%20Space%20v4%20Briefing.pdf},
}

@misc{kepler_k2,
  author       = {Carney, Stephen},
  title        = {Kepler / K2},
  howpublished = {National Aeronautics and Space Administration},
  year         = {2025},
  url          = {https://science.nasa.gov/mission/kepler/},
}

@misc{deep_space_1,
  author       = {Lindsey, Lauren},
  title        = {Deep Space 1},
  howpublished = {National Aeronautics and Space Administration},
  year         = {2024},
  url          = {https://science.nasa.gov/mission/deep-space-1/},
}

@misc{HROV_nereus,
  author       = {Hadal Ecosystems Studies Program},
  title        = {HROV Nereus},
  howpublished = {Hadal Ecosystems Studies},
  year         = {2014},
  url          = {https://web.whoi.edu/hades/nereus/},
}

\end{document}